\documentclass[conference]{IEEEtran}
\IEEEoverridecommandlockouts
\usepackage{cite}
\usepackage[utf8]{inputenc}
\usepackage{amsmath,amssymb,amsfonts}
\usepackage{verbatim}
\usepackage[ruled]{algorithm2e}
\usepackage{graphicx}
\usepackage{textcomp}
\usepackage{stfloats}
\usepackage{xcolor}
\usepackage{tikz}


\def\BibTeX{{\rm B\kern-.05em{\sc i\kern-.025em b}\kern-.08em
    T\kern-.1667em\lower.7ex\hbox{E}\kern-.125emX}}
\begin{document}

\title{Detection of AI-Synthesized Speech Using Cepstral \& Bispectral Statistics\\
}

\author{\IEEEauthorblockN{ Arun Kumar Singh}
\IEEEauthorblockA{Indian Institute of Technology, Jammu\\
Jammu J\&K,  India \\
2019PCS0017@iitjammu.ac.in}

\and
\IEEEauthorblockN{Priyanka Singh}
\IEEEauthorblockA{ Dhirubhai Ambani Institute of\\ Information and Communication Technology\\
Gandhinagar, Gujarat,  India \\
priyanka\_singh@daiict.ac.in}
}
\maketitle
\begin{abstract}

Digital technology has made possible unimaginable applications come true. It seems exciting to have a handful of tools for easy editing and manipulation, but it raises alarming concerns that can propagate as speech clones, duplicates, or maybe deep fakes. Validating the authenticity of a speech is one of the primary problems of digital audio forensics. We propose an approach to distinguish human speech from AI synthesized speech exploiting the Bi-spectral and Cepstral analysis. Higher-order statistics have less correlation for human speech in comparison to a synthesized speech. Also, Cepstral analysis revealed a durable power component in human speech that is missing for a synthesized speech. We integrate both these analyses and propose a model to detect AI synthesized speech.
\\
\end{abstract}


\section{Introduction}
In today's era, advances in Artificial Intelligence and Deep Neural Networks have led to very significant results in creating a more realistic type of synthesized audio and speech \cite{b2,b4}. Speech cloning and duplication via training the neural networks using powerful AI algorithms lead to synthesized speech.   However, these advancements had also resulted in many misuses of this technology. Many dangerous fakes have been created, which can harm in many possible ways. Hence,  the authenticity of digital data is the prime concern for all of us.
\

The field of speech forensics has progressed a lot over the past few decades, but very few schemes have proposed detecting AI synthesized content and speech. Some techniques address speech spoofing \cite{b6} and tampering, but they are not explicit for detecting AI synthesized speech. Hany discussed how the forgery of a signal affects the correlations of the higher-order statistics \cite{b1} but not discussed AI synthesized content. \ 

A comparison between the different features for synthetic speech, spectral magnitudes, and phase of the statistical measures presented in \cite{b3} to distinguish the human speech from the AI synthesized speech. First-order Fourier coefficients or second-order power spectrum correlations can be easily tuned to match the human speech while synthesizing speech. On the other hand, third-order bispectrum correlations are hard to adjust and can discriminate between human and AI speech \cite{b7}. One such work was proposed in \cite{b7} where they used bicoherence magnitude and phase to distinguish human speech from AI speech collected from various sources like Google AI, Siri, Baidu etc. Muda et al. \cite{b8} presented the distinction between male and female speeches using Mel Frequency Cepstral Coefficient (MFCC).  MFCC are useful features to identify vocal tracts. Synthetic speech detection using the temporal modulation technique presented in \cite{b9} also used MFCC as one of their features. However, they did not include two primary features related to the MFCC: $\Delta$-Cepstral and $\Delta^{2}$-Cepstral.

The motivation of the proposed algorithm comes from the urgent need of more robust features for discriminating AI synthesized speech from human speech. In the proposed work, we have considered synthesizers that are not too famous but readily available to common users. We combine multiple primary features, specifically higher-order spectral correlations revealed by Bispectral analysis, and MFCC, $\Delta$-Cepstral and $\Delta^{2}$-Cepstral revealed by Mel Cepstral analysis, to account for the enhanced discrimination between human speech and  AI synthesized speech. Bispectral analysis can identify components in AI synthesized speech that are absent in human speech \cite{b7}. The process of AI synthesis may induce some correlations in the AI synthesized speech  as it passes through different layers of neural network. These correlations are hard to remove as they are likely to be generated due to the fundamental properties of the synthesis process \cite{b1}. These correlations are lacking in the recorded human speech. Mel Cepstral analysis of the speech reveals strong power components in human speech, which is not present in the AI synthesized speech. These power components may be present in human speech due to the vocal tract, which in contrast, is not the case with AI-Synthesised speech. We integrate Bispectral analysis and Mel Cepstral analysis and perform various experiments to test the efficacy for classifying the Human and AI speech. The audio data we accounted for, in our experiments, is uncleaned audio data which may contain noise and interference as we get in real life scenarios.

The rest of the paper is organized as follows. Section \ref{sec:Methods} gives a brief overview of the key concepts used in the proposed algorithm, section \ref{sec:Classi} provides the details of the dataset, classification model and parameters used. Section \ref{sec:Results} discusses the result findings of the proposed method.

\section{Preliminaries}\label{sec:Methods} 
In this section, a brief overview of the higher-order statistics is given that we have used in our proposed algorithm as distinguishing features. Analysis of Mel Frequency Cepstral Coefficient (MFCC) and visualization using the Mel spectrogram is described. Delta and Delta Square related to Mel Cepstrum is briefed.

\subsection{Bispectral Analysis}

The bispectrum of the signal represents a higher-order correlation in the Fourier domain. The simple Fourier coefficients represent the first-order correlation or first-order statistics. \  

An audio signal $y(k)$ is decomposed into different frequencies according to the Fourier transform :
\begin{equation}
Y(\omega) = \sum_{k=-\infty}^{\infty} y(k).e^{-ik\omega}
\end{equation}
with $\omega\ \epsilon\ [-\pi,\pi]$. Power spectrum of the signal $P(\omega)$ is generally used to detect second order correlations, given by : $P(\omega) = Y(\omega).Y^{*}(\omega)$ where $*$ denotes the complex conjugate. Power spectrum is blind to higher order correlations that means we cannot detect higher order correlations using power spectrum. However, these higher order correlations can be detected using bispectral analysis. We find bispectrum of the signal to calculate third order correlations which is given by:
\begin{equation}
B(\omega_{1},\omega_{2}) = Y(\omega_{1}).Y(\omega_{2}).Y^{*}(\omega_{1} + \omega_{2})
\end{equation}
Unlike the power spectrum, the bispectrum in the Equation (3) is a complex valued quantity. So for the purpose of simplicity and interpretation for our problem, it is suitable to represent or use the complex bispectrum with respect to it's magnitude : 
\begin{equation}
\mid B(\omega_{1},\omega_{2}) \mid\ =\ \mid Y(\omega_{1})\mid .\mid Y(\omega_{2})\mid .\mid Y(\omega_{1} + \omega_{2})\mid
\end{equation}
and Phase :
\begin{equation}
\angle B(\omega_{1},\omega_{2}) = \angle Y(\omega_{1}) + \angle Y(\omega_{2}) - \angle Y(\omega_{1} + \omega_{2})
\end{equation}
Also for the purpose of scaling and simplicity in calculations, it will be helpful to use the normalized bispectrum \cite{b5}, the bicoherence : 
\begin{equation}
B_{c}(\omega_{1},\omega_{2}) = \frac{Y(\omega_{1}).Y(\omega_{2}).Y^{*}(\omega_{1} + \omega_{2})}{\sqrt{\mid Y(\omega_{1}).Y(\omega_{2})\mid^{2}.\mid Y(\omega_{1} + \omega_{2})\mid^{2}}}
\end{equation}
This normalized bispectrum yields magnitude in the range $[0,1]$. But we have used the other normalized process for bispectral magnitude and phase which also yields the range into $[0,1]$. \ 

For the purpose of efficient calculation, we divide each speech samples of length $N$ into approximately $K \approx 100$ smaller samples of length $N/K$. The bispectral magnitude and phase of these $K$ segments are summed over different $\omega$ values and average value is taken. 
\begin{equation}
 \mid \widehat B(\omega_{1},\omega_{2}) \mid\ =\ \frac{1}{K} \sum_{K}( \mid Y_{K}(\omega_{1})\mid\mid Y_{K}(\omega_{2})\mid\mid Y_{K}(\omega_{1} + \omega_{2})\mid) 
\end{equation}
\begin{equation}
\angle \widehat B(\omega_{1},\omega_{2}) =  \frac{1}{K} \sum_{K}(\angle Y_{K}(\omega_{1}) + \angle Y_{K}(\omega_{2}) - \angle Y_{K}(\omega_{1} + \omega_{2})) 
\end{equation}

\subsection{Mel Frequency Cepstral Coefficient (MFCC) and Analysis}

The speech generated by humans can be uniquely identified due to the vocal tract's shape that includes human oral organs, during the speaking. In general, these different vocal tract shapes helps in determining the type of sound that we speak. The Mel Frequency Cepstral represents the short-time power spectrum of the audio. It can be used to filter speech based on vocal tract shape. This is represented using MFCCs. On a similar hypothesis, there will be differences in MFCCs values of Human speech and AI-synthesized speech as the speech generated by AI are not generated from vocal tracts.

 For our study,  we have considered four types of speech. The first one is human speech, and the other three are AI synthesized speech from three different sources, namely Spik.AI, Natural Reader, and Replica AI. These three kinds of AI speech are generated from different text to speech-generating AI engines.  Mel spectrogram for the four types of speeches is represented in Figure \ref{fig2}.
\  

The MFCCs are calculated from the magnitude spectrum of short term Fourier transform of the audio signal. The short term Fourier transform of the audio signal $y(k)$ is given by :
\begin{equation}
Y(\omega) =\ \mid Y(\omega) \mid e^{\ j\phi(\omega)}
\end{equation}
where $\mid Y(\omega) \mid$ is magnitude spectrum and $\phi(\omega)$ is the phase spectrum. For calculating MFCC, the entire speech is split into overlapping segments called windows. After that, a Fourier transform is performed for each segment, which is used to derive the power spectrum. Mel Frequency Filter is applied to the power spectrum obtained, and then discrete cosine transform (DCT) of the Mel log power is taken. The MFCCs represents the amplitude of the obtained spectrum after DCT is performed.\

Other parameters associated with MFCC useful as a feature for the distinction of speech are $\Delta$-Cepstrum and $\Delta^{2}$-Cepstrum. Change in MFCC coefficients is given by $\Delta$-Cepstrum, i.e., $\Delta$-Cepstrum is the difference between the current MFCC coefficient and the previous MFCC coefficient. Similarly, Change in $\Delta$-Cepstrum values is given by $\Delta^{2}$-Cepstrum, i.e., $\Delta^{2}$-Cepstrum is the difference between current $\Delta$-Cepstrum value and the previous $\Delta$-Cepstrum value. All these three parameters act as strong 
traits for representing Cepstral Analysis.

\begin{figure}[ht]
  \centering
  \framebox[3.5in]{\includegraphics[width=0.45\textwidth,height=14cm]{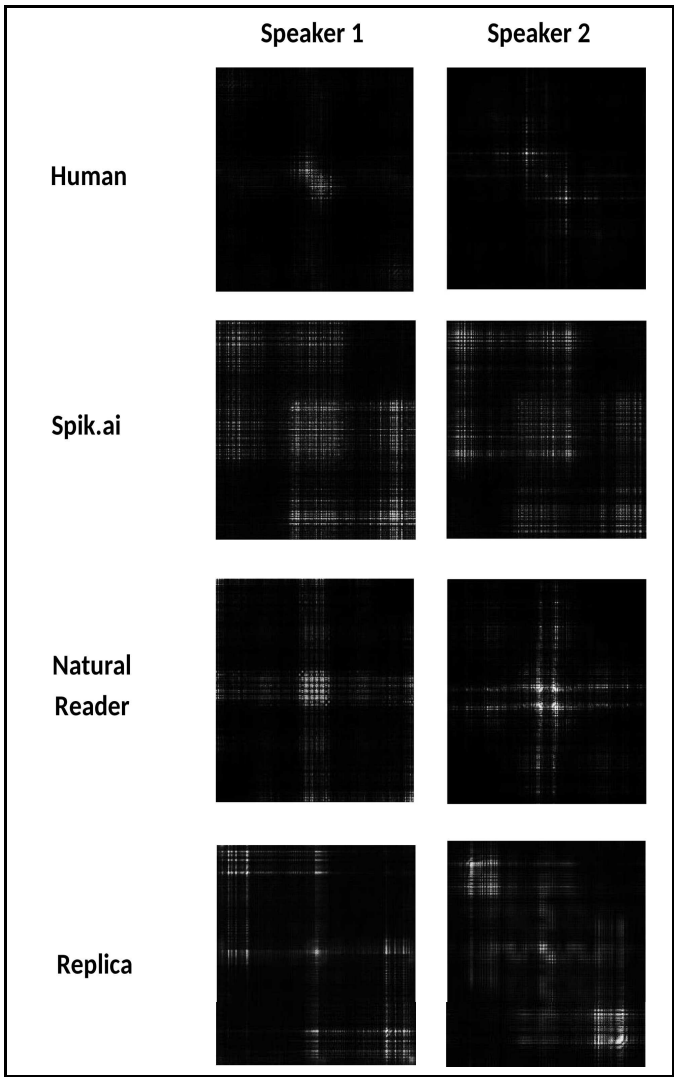}}
  \caption{Bicoherent magnitude of two speakers for human and three synthesized speech. The magnitude plot is shown on an intensity of the scale $[0,1]$}
  \label{fig1}
\end{figure}

\begin{figure*}[ht]
\centering
\setlength\fboxsep{0.1pt}
\setlength\fboxrule{0.15pt}
\framebox[7.1in]{\includegraphics[width=13.95cm, height=18cm,keepaspectratio]{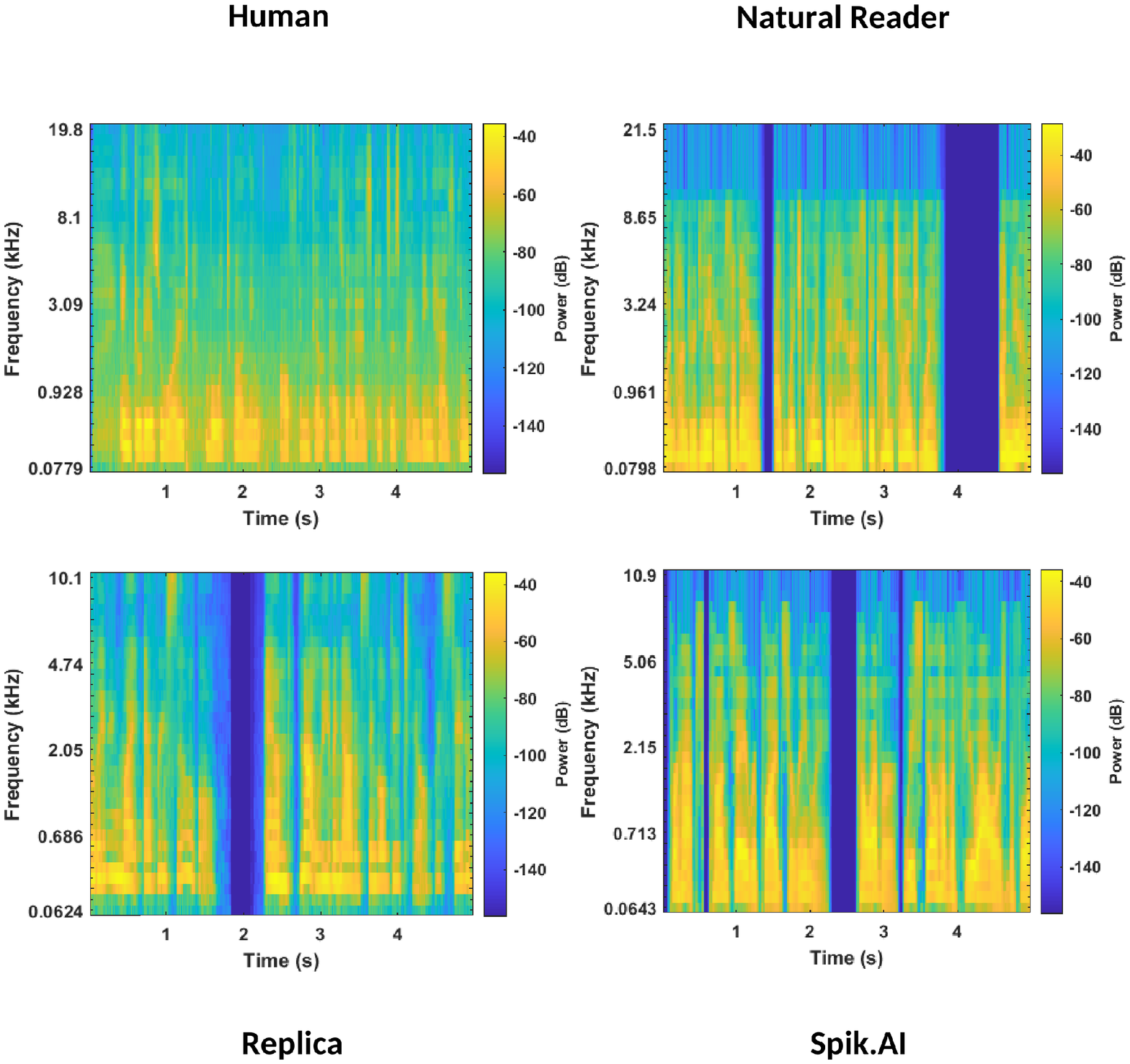}}
\caption{Melspectrogram for different four types of speeches}
\label{fig2}
\end{figure*}

\section{Experiments and Analysis}
\label{sec:Classi} 

In this section, we have described the data set created by us by collecting speech samples from different sources. Some relics for the observations, the classification model, and the parameters used in the study are also described.

\subsection{Data Set}
We collected speech samples from different sources and created a data set for our experiment. Our data set comprises of total 704 different samples.  For human speech, we have a total of 250 samples, where 110 speech samples are taken from the Kaggle data repository, and 140 speech samples are recorded with the microphone. For diversity in our data set, we took human speech samples from both male and female speeches. For AI synthesized speech, we have collected samples from three different sources: Natural Reader, Spik.AI, and Replica. We have taken 79 samples from Natural Reader,  230 samples from Spik.AI, and 126 samples from Replica AI. These AI synthesized speech engines are based on text-to-speech synthesis. We collected the sample from these not too famous synthesizers because, the AI advancement has increased to so much extent that these types of unnoticed sources has increased a lot and readily available for use. And the old techniques were limited to the testing over some famous synthesizers only. Before processing each speech sample (both AI synthesized and human), it is trimmed to average of 5 secs each.  Since audios exist in two channels, i.e., mono and stereo, we converted all speech samples into monotype for the equal basis of comparison.  These samples were then distributed randomly into 622 training and 82 testing data. Also we didn't pre-process our data so the noise and interference are readily available. The reason behind to use unprocessed data is that, in real life applications we don't get processed data and we want to test how much our features (independently and combined) can be effective over unprocessed audio data.

\subsection{Relics for the Observations}
We demonstrated some of the preliminary results in the form of images in Figure \ref{fig1}, showing the different patterns achieved after processing different speeches. The bicoherence magnitude for two speakers of each category is represented in Figure \ref{fig1}. The first row represents the relic of human speech. Similarly, second, third, and fourth row represent relics from AI synthesized speech, i.e., Spik.AI, Natural Reader, and Replica AI. Similarly, the columns represent the normalized bispectral magnitude for speaker 1  and speaker 2 for the corresponding speeches. All images are shown on the same intensity scale.\

On observing the relics, we can see a glaring difference between the magnitude of human speech and all other AI synthesized speech.  These variations can be due to significant spectral correlations present in AI synthesized speech but absent in human speech. These spectral correlations in  AI synthesized speech are induced due to neural network architecture. In particular, the long-range temporal connections between the layers of neural networks may be the cause.\ 

Four Mel spectrograms of different four speeches, i.e., Spik.AI, Replica, Natural Reader, and Human speech is shown in  Figure \ref{fig2}.  Mel power scale is given on the right of each spectrogram. From reference from scale we can see that dark blue colour indicate weak power component and yellow colour represents strong power component in the spectrogram \ref{fig2}. It is observed that a strong power component is missing in all types of AI-synthesized speech, which is not in the case of human speech. It may be due to the absence of vocal tract during the generation of  AI-synthesized speech. These differences shown in the Mel spectrogram encourage us to add MFCC as a feature for discriminating speeches.

\subsection{Classification Models, Parameters, Scenarios and Test Cases}

The bicoherence magnitude and phase are calculated, as mentioned in Equation 6 and Equation 7, respectively, for all human speech samples and three types of AI synthesized speech. These quantities are then normalized to the range [0,1] and then used to calculate the machine learning model's higher-order statistical parameters. 

We calculate the first four statistical moments for both magnitude and phase. Let M and P be the random variables denoting the underlying distribution of bicoherence magnitude and phase. The first four statistical moments are given by:\\
\begin{itemize}
    \item Mean , $\mu_{X} = E_{X}[X]$ \\
    \item Variance , $\sigma_{X} = E_{X}[(X-\mu_{X})^{2}]$\\
    \item Skewness , $\gamma_{X} = E_{X}[(\frac{X-\mu_{X}}{\sigma_{X}})^{3}]$\\
    \item Kurtosis , $\kappa_{X} = E_{X}[(\frac{X-\mu_{X}}{\sigma_{X}})^{4}]$\\
\end{itemize}
where $E_{X}[.]$ is the expected value operator with regards to random variable X. For the magnitude, we represent X = M, and for the phase X = P, these four moments are calculated by replacing this expected value operator with average. Also, for each speech sample, the mean and variance of MFFC, $\Delta$-Cepstrum, and $\Delta^{2}$-Cepstrum are calculated. It contributes to a 15-D feature vector for each speech sample. The first 8 entries represent the four moments for magnitude and phase. The next 6 entries represent the mean and variance of MFCC, $ \Delta $-Cepstrum, and $\Delta^{2}$-Cepstrum and last entry represent the class of speech, i.e., Human, Natural Reader, Spik.AI or Replica.

\begin{table*}[!ht]
\centering
\caption{Area Under the Curve (AUC) Values in ROC for different ML classifiers.}
\label{tab:AUC_values_AllExperiments}
\resizebox{\textwidth}{!}{%
\begin{tabular}{|c|c|c|c|c|}
\hline
                       & \multicolumn{2}{c|}{\textbf{Multi-Class}} & \multicolumn{2}{c|}{\textbf{Binary Class}} \\ \hline
\textbf{Various Models} &
  \begin{tabular}[c]{@{}c@{}}Bicoherence \\ Magnitude \& Phase\end{tabular} &
  \begin{tabular}[c]{@{}c@{}}Bicoherence \\ Magnitude \& Phase\\ +\\ MFCC \& Delta Cepstral\\ +\\ Delta Square Cepstral\end{tabular} &
  \begin{tabular}[c]{@{}c@{}}Bicoherence \\ Magnitude \& Phase\end{tabular} &
  \begin{tabular}[c]{@{}c@{}}Bicoherence \\ Magnitude \& Phase\\ +\\ MFCC \& Delta Cepstral\\ +\\ Delta Square Cepstral\end{tabular} \\ \hline
Linear Discriminiant   & 0.81                & 0.99                & 0.79                 & 0.99                \\ \hline
Quadratic Discriminant & 0.63                & 0.98                & 0.71                 & 0.98                \\ \hline
Linear SVM             & 0.80                & 0.99                & 0.80                 & 0.98                \\ \hline
Quadratic SVM          & \textbf{0.83}       & \textbf{0.99}       & \textbf{0.84}        & \textbf{0.99}       \\ \hline
Weighted KNN           & 0.82                & 0.99                & 0.82                 & 0.99                \\ \hline
Boosted Trees Ensemble & 0.84                & 0.98                & 0.84                 & 0.96                \\ \hline
Logistic Regression    & --                  & --                  & 0.79                 & 0.99                \\ \hline
\end{tabular}%
}
\end{table*}

We perform experiments considering following scenarios:
\begin{itemize}
   
    \item  Scenario 1: In this setup, we classify speech samples into two classes i.e., Human vs. AI-synthesized (Natural Reader, Spik.AI or Replica). It is a binary classification and main focus of this paper.
     \item Scenario 2: In this experiment, we classify speech samples into multiple classes i.e., Human, Natural Reader, Spik.AI and Replica.
\end{itemize}
\
The 15-D feature representation for these experiment scenarios differs only in the last entry. For scenario 1, the last entry can represent either of the two values, i.e., Human or AI synthesized. However, for scenario 2, it can represent any of the four classes, i.e., Human, Natural Reader, Spik.AI, or Replica. We perform machine learning-based classification for both the experiment scenarios with the feature mentioned earlier, with the intuition of different expected outcomes.

We calculated Area Under the Curve (AUC) for ROC of different classifiers for both the scenarios mentioned in Table \ref{tab:AUC_values_AllExperiments}. We did the experiments by using many different classifiers for the below two cases to justify how adding MFCC, $\Delta$-Cepstrum, and $\Delta^{2}$-Cepstrum into  third-order  bispectrum  correlations can act as a set of good features:
\begin{itemize}
    \item  Case 1: Considering only Bicoherence Magnitude and Phase as a feature for classification, similar to the experiment of \cite{b7}.
     \item Case 2: Adding MFCC, $\Delta$-Cepstrum, and $\Delta^{2}$-Cepstrum along with Bicoherence Magnitude and Phase.
\end{itemize}



By visualizing the data for both types of classification and based on our intuition, we tried a few of the learning algorithms to train the Machine Learning model for classifying samples. For binary classification, we did experiment with seven ML algorithms for training, i.e, Linear and Quadratic Discriminant, Linear and Quadratic SVM, Weighted KNN, Boosted Trees and Logistic Regression with 5-fold cross-validation. For multi-class classification, we did experiment with six ML algorithms for training, i.e., Linear and Quadratic Discriminant, Linear and Quadratic SVM, Weighted KNN and Boosted Trees with same 5-fold cross-validation.

\section{Results and Discussions}
\label{sec:Results} 

\begin{figure}
\centerline{\includegraphics[width=0.5\textwidth]{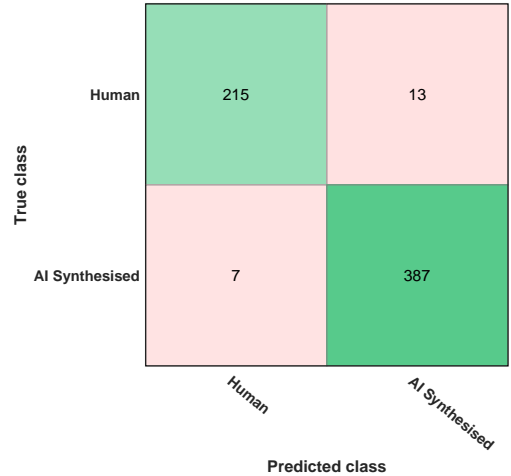}}
\caption{Confusion matrix for binary classification on training data set}
\label{fig6}
\end{figure}

\begin{figure}
\centerline{\includegraphics[width=0.5\textwidth]{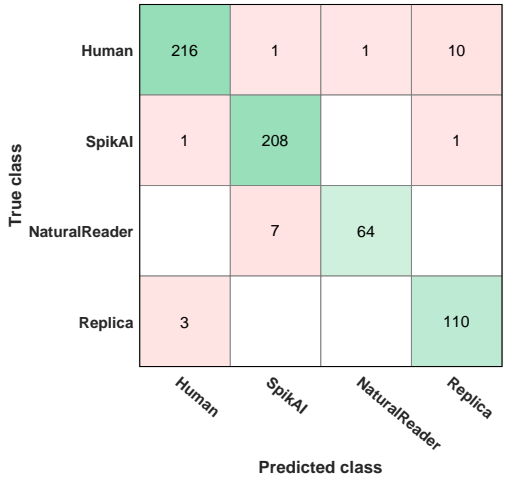}}
\caption{Confusion matrix for multi-class classification on training data set}
\label{fig5}
\end{figure}

After testing each classifier's performance by 5-fold cross-validation, we found that the binary classification Quadratic Support Vector Machine(Q-SVM) algorithm-based machine learning model has the highest accuracy with 96.3\%. For multi-class classification, also the Quadratic Support Vector Machine(Q-SVM) algorithm-based machine learning model has the highest accuracy with 96.1\%. We choose Quadratic SVM over all other classifiers though there was a marginal difference because the Kernel trick in SVM gives us an edge over all other classifiers. The Q-SVM model we used in our experiment had kernel function set to quadratic and kernel scale set to 2. During our experiments, we also observed that in a binary classification setting up kernel scale to 3 increases the training accuracy to 97.6\% and testing accuracy to 98.4\%. But that doesn't show a good result with multi-class classification. \\
Hence, for both binary class classification and multi-class classification, we chose Quadratic SVM as our classifier for constructing the trained ML model. The classifier's exact accuracy can be visualized with a confusion matrix, which indicates how many true values are falsely recognized. The confusion matrix for both the classifier Quadratic SVM binary class and Quadratic SVM multi-class is presented in Figure \ref{fig6} and Figure \ref{fig5}, respectively.

By observing both the confusion matrix, we  can see that binary classification results are more potent than the multi-class classification. In multi-class classification, we have false positives between Natural Reader and Spik.AI, Spik.AI and Replica, etc., which will be treated inside one class, i.e., AI synthesized in binary class classification. In the confusion matrix, we can observe that many of the Natural Reader speech samples are misclassified as Spik.AI samples. This may be due to their similarity in neural network engines. However, this thing does not matter much in the binary classification because all AI synthesized speech is classified as AI synthesized irrespective of what neural network architecture they follow. That is why binary classification between Human speech vs. AI synthesized speech has much better accuracy on cross-validation than multi-class classification.

\begin{figure}
\centerline{\includegraphics[width=0.5\textwidth]{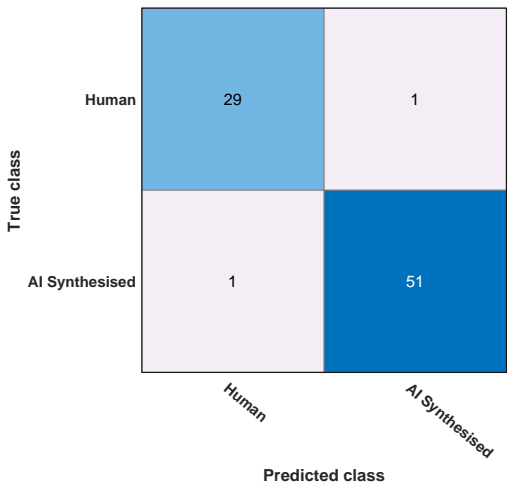}}
\caption{Confusion matrix for binary classification on test data set}
\label{fig7}
\end{figure}

\begin{figure}
\centerline{\includegraphics[width=0.5\textwidth]{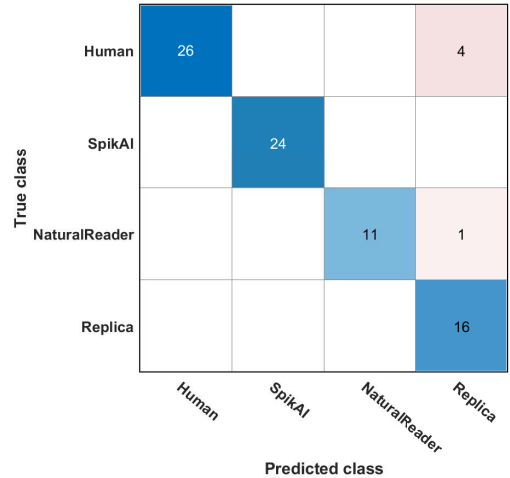}}
\caption{Confusion matrix for multi-class classification on test data set}
\label{fig8}
\end{figure}

After choosing our classifier and training the model, we tested our test data set over the trained model for the prediction. The test data consist of 82 random samples from all four types of speeches. Prediction is performed for both binary class classification and multi-class classification.  Figure \ref{fig7} and \ref{fig8} represents the confusion matrix for our final result of predicted data on binary classification and multiclass classification respectively.

The accuracy achieved on the prediction of test data on binary classification, which was our main motto, is 97.56\% with a miss classification rate of 2.43\%. For multi-class classification,we achieve accuracy of 93.9\%, with a miss classification rate of 6.09\%.  

We plan to study and integrate other discriminatory features to improve upon the accuracy and decrease the miss classification rate for our future work. Also, the scalability of the proposed model can be validated by testing with more massive datasets. More variants of experiment scenarios like classification based on gender, age, and accent can be done.

\end{document}